\pdfoutput=1
\documentclass[conference]{IEEEtran}
\IEEEoverridecommandlockouts
\usepackage{cite}
\usepackage{amsmath,amssymb,amsfonts}
\usepackage{algorithmic}
\usepackage{graphicx}
\usepackage{textcomp}
\usepackage{xcolor}

\usepackage{hyperref}
\usepackage{subcaption}
\usepackage{gensymb}

\def\BibTeX{{\rm B\kern-.05em{\sc i\kern-.025em b}\kern-.08em
    T\kern-.1667em\lower.7ex\hbox{E}\kern-.125emX}}
\begin{document}

\title{Multi-Task Learning with Loop Specific Attention for CDR Structure Prediction}

\author{\IEEEauthorblockN{Eleni Giovanoudi}
\IEEEauthorblockA{\textit{University of Thessaly} \\
Volos, Greece \\
egiovanoudi@uth.gr}
\and
\IEEEauthorblockN{Dimitrios Rafailidis}
\IEEEauthorblockA{\textit{University of Thessaly} \\
Volos, Greece \\
draf@uth.gr}
}

\maketitle

\begin{abstract}
The Complementarity Determining Region (CDR) structure prediction of loops in antibody engineering has gained a lot of attraction by researchers. When designing antibodies, a main challenge is to predict the CDR structure of the $H_3$ loop. Compared with the other CDR loops, that is the $H_1$ and $H_2$ loops, the CDR structure of the $H_3$ loop is more challenging due to its varying length and flexible structure. In this paper, we propose a Multi-task learning model with Loop Specific Attention, namely MLSA. In particular, to the best of our knowledge we are the first to jointly learn the three CDR loops, via a novel multi-task learning strategy. In addition, to account for the structural and functional similarities and differences of the three CDR loops, we propose a loop specific attention mechanism to control the influence of each CDR loop on the training of MLSA. Our experimental evaluation on widely used benchmark data shows that the proposed MLSA method significantly reduces the prediction error of the CDR structure of the $H_3$ loop, by at least 19\%, when compared with other baseline strategies. Finally, for reproduction purposes we make the implementation of MLSA publicly available at \url{https://anonymous.4open.science/r/MLSA-2442/}.
\end{abstract}


\section{Introduction}
\label{chap1}
The immune system is a complex network that defends the body against harmful infections. Essential elements of the immune system are antibodies, as they play a crucial role in recognizing and neutralizing these foreign pathogens \cite{maddur2020natural}. Their importance as therapeutic and diagnostic reagents has been known and exploited for almost a century \cite{maynard2000antibody}. However, the experimental determination of their structure is challenging, as it is both very expensive and time consuming \cite{kuroda2012computer}. In recent years, a promising solution to this problem seems to be antibody engineering, allowing the design of custom antibodies suitable for various biotechnological and biomedical applications \cite{chatenoud2014treatment}. Significant technological advances have accelerated and enhanced the discovery and development of antibody therapies, representing one of the fastest growing areas of the pharmaceutical industry \cite{kim2005antibody}.

\begin{figure}
\centering
\includegraphics[width=0.3\textwidth]{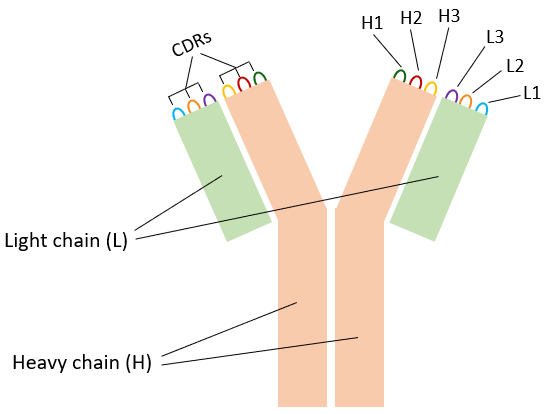}
\caption{The Y-shaped structure of an antibody \cite{fu2022antibody}.}
\label{Structure}
\end{figure}

Antibodies have a Y-shaped structure, formed by two identical heavy chains and two identical light chains, as illustrated in Figure \ref{Structure}~\cite{fu2022antibody}. Each of the two arms contains a variable region and a constant region \cite{stanfield2015antibody}. The variable region is responsible for recognizing and binding to a specific antigen through a set of six hypervariable loops, each of which forms a Complementarity Determining Region (CDR), located at the tip of the antibody's arms \cite{chiu2019antibody}. Structural modeling of CDRs is crucial since they adopt different conformations that determine the specificity of the antibody for its target antigen \cite{ling2020sagacity}. This way we are able to identify the amino acid residues in the CDRs that are likely to be involved in antigen binding. This information aids in the selection of antibodies for a particular application, or in the development of antibodies with improved binding properties \cite{saini2021bispecific}.

Five of these loops ($L_1–L_3$, $H_1$ and $H_2$) are structurally conserved, meaning that they tend to adopt a similar three-dimensional structure \cite{chothia1989conformations}, so state-of-the-art strategies predict them quite effectively \cite{north2011new}. However, the $H_3$ loop is typically longer and more flexible. Its length varies widely between different antibodies and its structure is more diverse than the other loops~\cite{shirai1996structural}. In addition, the $H_3$ loop depends on the chain orientation and multiple adjacent loops due to its placement between the heavy and light chains \cite{dunbar2013abangle}. Hence, the accurate prediction of the $H_3$ loop's structure remains a challenge \cite{almagro2014second}. Understanding the structure and function of the $H_3$ loop is an important area of research in antibody engineering and design. Several machine learning methods have been developed to address this issue, including deep generative models~\cite{luo2022antigen,shuai2021generative}, as well as, energy-based models~\cite{li2011vsgb,mintseris2007integrating} and graph-based strategies~\cite{kong2022conditional,shui2020heterogeneous}. However, as we will show later in Section~\ref{chap4}, baseline strategies have limited accuracy in the challenging case of predicting the CDR structure of the $H_3$ loop. What is missing from existing strategies, is that they try to predict the CDR structure of every type of loop, that is, the $H_1$, $H_2$ and $H_3$ loops, separately, and not jointly. 

Despite the unique characteristics of the $H_3$ loop, the three CDR loops $H_1, H_2,$ and $H_3$ share several structural and functional similarities. For example, the three CDR loops are all formed by amino acid residues and belong to the same heavy chain, located in its variable region \cite{chiu2019antibody}. Moreover, they have the same purpose, determining the specificity and affinity of the antibody for the antigen \cite{ling2020sagacity}. Finally, they are all subject to somatic hypermutation, a process that introduces random mutations into the DNA encoding the antibody gene segments, leading to the generation of a diverse repertoire of antibodies capable of recognizing a wide range of antigens \cite{clark2006trends}.
This means, that the three CDR loops could be co-learned simultaneously, accounting for the structural and functional similarities and differences.

Meanwhile, multi-task learning (MTL) is effective when tasks are related or have a similar underlying structure, as in the case of CDR loops. By jointly learning the tasks, a MTL model leverages the commonalities between them to improve its performance. This approach has numerous applications~\cite{ruder2017overview}, such as in Natural Language Processing \cite{collobert2008unified,liu2016deep}, Speech Recognition \cite{deng2013new,kim2017joint}, Computer Vision \cite{gao2019nddr,girshick2015fast} and Drug Discovery \cite{lin2022generalizeddta,liu2022structured}. However, so far, there have been no attempts to apply the concept of MTL to the problem of predicting the structure of CDRs.

An interesting approach is presented by Jin et al. \cite{jin2021iterative}, introducing a graph neural network to learn a representation of the antibody sequence and structure. More specifically, the nodes of the graph indicate amino acid residues and the edges reflect spatial interactions between residues. As the model unravels the sequence, the predicted structure is iteratively refined. Nonetheless this method learns the structure of each CDR loop separately, in the context of single-task learning. To overcome the shortcomings of existing strategies, in this study we present the MLSA method making the following contributions:
\begin{itemize}
    \item We design a novel architecture, that exploits the information from the $H_1$ and $H_2$ loops to improve the prediction of the challenging $H_3$ loop. Considering the CDR loops of the heavy chain, $H_1, H_2,$ and $H_3$, as three different tasks, the MLSA model is jointly trained on them.
    \item In addition, we propose a loop specific attention mechanism, to properly weigh the importance of each CDR loop in the joint learning of the $H_1$, $H_2$ and $H_3$ loops. In doing so, we consider the structural and functional similarities and differences between the three CDR loops during the training of the MLSA model.
\end{itemize}

The rest of the paper is organized as follows, Section \ref{chap2} provides an overview of related work, and Section \ref{chap3} describes the architecture of the proposed MLSA model. In Section \ref{chap4}, we present our experimental evaluation, and Section \ref{chap5} concludes our work.

\section{Related Work}
\label{chap2}

Deep learning has proven to be an efficient approach in molecular modeling due to its ability to learn complex relationships between molecular structures and properties directly from raw data \cite{bule2021rise}. One of the most popular methods in the field of CDR structure prediction comes from Jumper et al. \cite{jumper2021highly} and is based on multiple sequence alignment of homologous sequences with the use of equivariant neural networks and an attention mechanism.

Deep Generative Models (DGMs) are an important area of deep learning. The main idea is to learn the underlying probability distribution of the data \cite{arvanitidis2017latent}. For example, Luo et al.~\cite{luo2022antigen} combine diffusion-based generative models with equivariant neural networks. This way, they co-design sequences and structures for specific antigens. Shuai et al. \cite{shuai2021generative} create synthetic libraries of antibody sequences using a deep generative language model. To design antibodies, they formulate the CDR prediction task as an autoregressive sequence generation problem and filling in missing sections of the sequence with natural language text. Another approach is based on a Variational Autoencoder architecture~\cite{lai2022end}. The proposed architecture generates structures across a broad range of fold space, making it a universal model for protein structure prediction. Despite the capabilities of DGMs, permutation invariance can be a challenge. Most of DGMs are inherently permutation sensitive, meaning that they may assign different probabilities to different permutations of the same graph or molecule \cite{liu2021graphebm}.

Energy-based models are also used in molecular modeling, as they provide a quantitative framework for describing the interactions and energetics of molecular structures. An energy model assigns energy value to each possible conformation, with lower energies given to data points closest to the real data \cite{li2011vsgb}. Using Monte Carlo simulation or other techniques, an energy-based model reaches a local energy minimum, which represents the most stable and likely conformations of the protein \cite{welling2011bayesian}. For instance, Norn et al.~\cite{norn2017high} break the antibody structure into fragments and introduce a Monte Carlo sampling algorithm. This way, their model searches for low-energy combinations of the fragments that can form a complete antibody structure. Mintseris et al.~\cite{mintseris2007integrating} sample different conformations of the protein complexes and use statistical pair potentials to evaluate their energetics and predict their structure. Ruffolo et al.~\cite{ruffolo2022antibody} generate initial conformations via a deep residual convolutional neural network and then use a physics-based energy minimization method to further refine the structures and improve the accuracy. Fu and Sun \cite{fu2022antibody} propose a constrained energy model that considers both the energetics of the CDR structure and the constraints imposed by the antigen binding site. This method guides the optimization process towards conformations that have the lowest energy and best binding interactions with the antigen. Generally, energy-based models are able to capture the underlying physical principles that govern protein folding and binding. However, they can also be computationally expensive \cite{ingraham2019generative}.

In addition, graph-based models have been widely used in molecular modeling due to their ability to capture the structure and interactions of molecules \cite{shui2020heterogeneous}. Ruffolo et al. \cite{ruffolo2023fast} employ a pre-trained language model followed by graph networks to predict the coordinates. Kong et al.~\cite{kong2022conditional} formulate the antibody design problem as a conditional graph translation task, where the target antigen and the light chain of the antibody are incorporated as additional components. Abanades et al.~\cite{abanades2022ablooper} represent the sequence as a graph using equivariant graph neural networks. The proposed model simultaneously predicts multiple structures for each loop and compares them among themselves to identify the most representative conformations. Overall, graph-based models are a valuable tool for molecular modeling, as they offer a powerful and flexible framework \cite{kearnes2016molecular}. Nonetheless, such models ignore the structural and functional similarities between the three CDR loops, as learning is performed in individual graphs and do not follow a joint strategy.

To accelerate the process of identifying and producing new drugs, MTL has gained popularity as a drug discovery and development strategy~\cite{elbadawi2021advanced}. For example, in the case of personalized therapy based on tumor molecular features, MTL across different drugs significantly enhances the efficiency and interpretability of drug prediction models \cite{yuan2016multitask}. In treatment trials for HIV, a combination of drugs is typically used. Thus, MTL is performed to predict the outcome of a combination for a specific strain of HIV \cite{bickel2008multi}. Moreover, accounting for the fact that multi-target drugs have significant advantages over single-target drugs in the treatment of complex diseases, MTL can leverage the commonality between multiple targets to share knowledge among target specific QSAR models \cite{rosenbaum2013inferring}. The main difference between our work and state-of-art strategies for drug discovery is that we follow a MTL strategy to improve the accuracy of 3D structure of the challenging $H_3$ loop in antibody engineering.

\section{The Proposed MLSA Model}
\label{chap3}

\subsection{Overview
}
\begin{figure*}
\centering
\includegraphics[width=\textwidth]{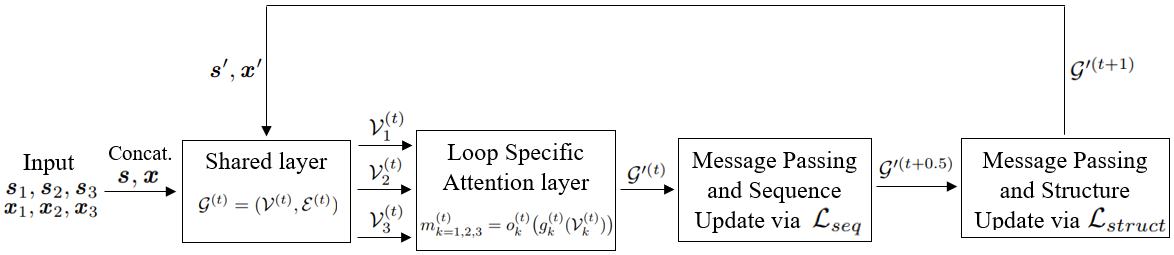}
\caption{The architecture of the proposed MLSA model.}
\label{Architecture}
\end{figure*}

The multi-task learning architecture of the proposed MLSA model is designed to predict the 3D structure of CDRs, leveraging a combination of shared and loop specific layers. The core idea is to present an antibody as a graph, which encodes both its sequence and 3D coordinates \cite{jin2021iterative}, and apply attention to each loop separately \cite{liu2019end}. This way the model is able to focus on task-related features and employ the valuable information from the $H_1$ and $H_2$ loops in the learning of $H_3$. A brief demonstration of the MLSA architecture is illustrated in Figure \ref{Architecture}. 

In particular, the input is the sequences $\textbf{\textit{s}}_1, \textbf{\textit{s}}_2, \textbf{\textit{s}}_3$ and the coordinates $\textbf{\textit{x}}_1, \textbf{\textit{x}}_2, \textbf{\textit{x}}_3$ corresponding to the three loops $H_1, H_2, H_3$, respectively. Initially, we combine them into a unified $\textbf{\textit{s}}$ and $\textbf{\textit{x}}$ via concatenation. This information is then used to construct a graph $\mathcal{G}^{(t)}(\textbf{\textit{s}}) = (\mathcal{V}^{(t)}, \mathcal{E}^{(t)})$, where $\mathcal{V}^{(t)}$ represents the node features in the $t^{th}$ iteration and $\mathcal{E}^{(t)}$ represents the edge features. Next, we split $\mathcal{V}^{(t)}$ into three subsets $\mathcal{V}^{(t)}_1, \mathcal{V}^{(t)}_2, \mathcal{V}^{(t)}_3$, one for each loop, and forward them to the loop specific attention layer separately. Thus, we extract the attention masks $m^{(t)}_1, m^{(t)}_2,$ and $m^{(t)}_3$, which are concatenated into the new set of node features. In doing so, each attention mask automatically determines the importance of the respective loop for the shared representation, and we therefore balance the influence of each loop on the joint learning of MLSA, accordingly. This leads to a new graph $\mathcal{G}'^{(t)} = (\mathcal{V}'^{(t)}, \mathcal{E}'^{(t)})$, which is used to update the sequence and then the structure. For the sequence we use the loss function $\mathcal{L}_{seq} = \sum_t \mathcal{L}_{ce}(\textbf{\textit{s}}'_t, \textbf{\textit{s}}^{ground}_t)$, which measures the discrepancy between the predicted and true residue types as the cross entropy loss $\mathcal{L}_{ce}$ \cite{jin2021iterative}. For the structure we use the loss function $\mathcal{L}_{struct} = \sum_t \mathcal{L}_d^{(t)} + \mathcal{L}_{\beta}^{(t)} + \mathcal{L}_{ca}^{(t)}$, which is the sum of three constitute losses. The distance loss $\mathcal{L}_d$ measures the difference between the predicted and true distances of the alpha carbons using the Huber loss \cite{Huber}. The dihedral angle loss $\mathcal{L}_{\beta}$ calculates the mean square error between the predicted and the actual dihedral angles formed by the atoms \cite{xia2021geometric}. The $C_a$ angle loss $\mathcal{L}_{ca}$ computes the difference between the predicted and true angles defined by the alpha carbons. At the end of each iteration, we return $\textbf{\textit{s}}'$ and $\textbf{\textit{x}}'$ to the shared layer for the next iteration, until the optimization algorithm converges. This means that the output of the proposed MLSA model is $\textbf{\textit{x}}'$, which consists of $\textbf{\textit{x}}'_{k, i}$ for the $i^{th}$ residue of the predicted coordinates $\textbf{\textit{x}}'_k$ of each loop $k=1,2,3$.

\subsection{Shared layer}
To exploit the information from the three loops simultaneously, we first obtain the sequences and coordinates corresponding to these regions for each antibody. Each sequence $\textbf{\textit{s}}_k$, is a series of residues $\textbf{\textit{s}}_{k,1}\textbf{\textit{s}}_{k,2} \cdots \textbf{\textit{s}}_{k,p}$, where each $\textbf{\textit{s}}_{k,i}$ represents one of the 20 known amino acids. For each amino acid, we extract the 3D coordinates of three basic atoms, the alpha carbon (a), the carbon (c), and the nitrogen (n). Consequently, on residue level, each one of the $\textbf{\textit{x}}_k$ coordinates can be further subdivided into $\textbf{\textit{x}}_{k,i,a}, \textbf{\textit{x}}_{k,i,c},$ and $\textbf{\textit{x}}_{k,i,n}$, respectively. We combine the loops into a unified form via concatenation, that is the sequences into $\textbf{\textit{s}}$ and the coordinates into $\textbf{\textit{x}}$, and forward them to the shared layer.

We then construct a graph $\mathcal{G}(\textbf{\textit{s}}) = (\mathcal{V, E})$ illustrating the structure of the antibody and the interactions between the atoms. $\mathcal{V} = \{\textbf{\textit{v}}_1, \cdots, \textbf{\textit{v}}_r\}$ is the set of the node features, that is the amino acid residues. More specifically, because of the three aforementioned atoms a, c, n, each $\textbf{\textit{v}}_i$ consists of three dihedral angles $(\phi_i, \psi_i, \omega_i)$. $\mathcal{E} = \{\textbf{\textit{e}}_{ij}\}_{i\neq j}$ are the edge features expressing the spatial interactions between the residues. These relations are determined by distance and orientation \cite{ingraham2019generative}.  

\subsection{Loop Specific Attention layer}

The attention layer is designed to selectively focus on task-related features. Therefore, the model consists of three task-specific attention modules with the same design, one for each loop. The set of node features $\mathcal{V}$ is divided into three subsets $\mathcal{V}_k$, each of which contains the information for the corresponding loop and is forwarded to the respective attention module. Each module applies a soft attention mask by focusing on the most important parts of the input. We denote the learned attention mask in the $t^{th}$ iteration for loop $k$ as $m_k^{(t)}$~\cite{liu2019end}, which is calculated as follows:
\begin{equation}
    m_k^{(t)} = o_k^{(t)}\bigl( g_k^{(t)}(\mathcal{V}_k^{(t)})\bigl).
\end{equation} 
The terms $o_k^{(t)}$ and $g_k^{(t)}$ are a pair of convolutional layers\footnote{In our implementation, we tune $q$ convolutional layers.}, composed of [$z$ × $z$] kernels, with batch
normalization, following a ReLU and a Sigmoid activation function respectively. Both $o_k^{(t)}$ and $g_k^{(t)}$ present the $k^{th}$ task-specific attention mask in iteration $t$. The attention mask is learned in a self-supervised fashion and sigmoid ensures that $m_k^{(t)} \in [0,1]$. Hence, the masks represent the weights of the input elements, emphasizing those that are most important. After calculating all the attention masks, we concatenate them back together as the new node features $\mathcal{V}'^{(t)} = \{m_1^{(t)};m_2^{(t)};m_3^{(t)}\}$. This results in a new graph $\mathcal{G}'^{(t)}$ with modified node features.

\subsection{Optimization}

In each iteration $t$, we apply attention to the node features, and more specifically to each $\mathcal{V}_k^{(t)}$ separately. This way, we obtain three attention masks $m_k^{(t)}$ which are then concatenated into a modified $\mathcal{V}'^{(t)}$, forming a new $\mathcal{G}'^{(t)}$. Next, we use a message passing network (MPN) to obtain the learned representations of the residues in the current graph $\mathcal{G}'^{(t)}$. We predict
the next residue of the sequence $\textbf{\textit{s}}'_{(t+1)} = softmax(\textbf{\textit{W}}_a\textbf{\textit{h}}^{(t)}_{t+1})$, where $\textbf{\textit{W}}_a$ is a learned weight matrix and $\textbf{\textit{h}}^{(t)}_{t+1}$ represents the residue $t+1$ in the graph $\mathcal{G}'^{(t)}$ \cite{jin2021iterative}. This leads to an intermediate graph $\mathcal{G}'^{(t+0.5)}$ which is encoded with a different MPN. Thus, we generate new coordinates for all residues $\textbf{\textit{x}}'^{(t+1)}_{i,f} = \textbf{\textit{W}}_x^f \textbf{\textit{h}}_i^{(t+0.5)}, \quad 1 \leq i \leq r, \quad f \in \{a,c,n\}$, where $\textbf{\textit{W}}_x^f$ is a learned weight matrix and $\textbf{\textit{h}}_i^{(t+0.5)}$ represents the residue $i$ in the graph $\mathcal{G}'^{(t+0.5)}$. As a result, a new graph $\mathcal{G}'^{(t+1)}$ is created for the next $t+1$ iteration, where both the sequence and the coordinates are updated. The loop continues until all residues have been predicted.

\section{Experiments}
\label{chap4}
\subsection{Dataset \& Evaluation Protocol}

\paragraph*{\textbf{Training and Testing Datasets}}
Following the evaluation protocol of state-of-the-art methods~\cite{abanades2022ablooper,jeliazkov2021robustification,ruffolo2022antibody}, training and testing is performed on different datasets of antibody structures. In particular, to train our model, we use a set of antibody structures from The Structural Antibody Database (SabDab)\footnote{\url{https://opig.stats.ox.ac.uk/webapps/newsabdab/sabdab/}}. SAbDab contains all the antibody structures available in the Protein Data Bank (PDB)\footnote{\url{https://www.rcsb.org/}}, annotated and presented in a uniform manner. Regarding the testing dataset, we use a separate benchmark dataset, that is the RosettaAntibody benchmark dataset including 49 targets \cite{marze2016improved,weitzner2017accurate}. In our experiments we use the publicly available training and testing datasets\footnote{\url{https://github.com/wengong-jin/RefineGNN/tree/main/data/sabdab}}, presented in~\cite{jin2021iterative}.

\paragraph*{\textbf{Training Phase}}
 Following~\cite{abanades2022ablooper,ruffolo2022antibody}, in the preprocessing phase, we retain only the structures with a resolution of 4Å and above, and eliminate overlapping sequences by setting a 99\% sequence identity cut off. This high threshold results from the fact that antibody sequences are conservative to a large extent and numerous therapeutic antibodies presented in databases are identical. In addition, we aim to familiarise the examined models with samples of minor alterations that cause structural variations. Notice that any targets included in the test set, or structures that shared 99\% similarity with a target's sequence, were deleted from the training dataset. Thus, a set of $1,467$ structures were produced for model training. Then, we randomly split this initial set into training and validation sets with an 8:2 ratio. This way, we obtained 1,186 structures for training and 281 for validation.

\paragraph*{\textbf{Testing Phase}}
The RosettaAntibody benchmark set contains 49 targets, with resolutions of 2.5Å or higher, selected from the PyIgClassify database \cite{adolf2015pyigclassify}. Additionally, the set was filtered to exclude antibodies with identical sequences in any of the heavy-chain CDR loops.

\paragraph*{\textbf{Evaluation}}
Following relevant studies~\cite{abanades2022ablooper,jin2021iterative,ruffolo2022antibody}, we report the Root Mean Square Deviation (RMSD) between the predicted and true structures which is defined as follows: 
\begin{equation}
    RMSD(\textbf{\textit{x}}^{ground}_k, \textbf{\textit{x}}'_k) = \sqrt{\frac{1}{n} \sum_{i=1}^n \parallel \textbf{\textit{x}}^{ground}_{k, i} - \textbf{\textit{x}}'_{k, i} \parallel^2},
\end{equation}
where $n$ is the number of points, $\textbf{\textit{x}}^{ground}_{k, i}$ is the $i^{th}$ residue of the ground truth coordinates $\textbf{\textit{x}}^{ground}_k$ of loop $k$, and $\textbf{\textit{x}}'_{k, i}$ is the $i^{th}$ residue of the predicted coordinates $\textbf{\textit{x}}'_k$ of loop $k$. Specifically, we use the Kabsch algorithm \cite{kabsch1976solution} to calculate RMSD based on the $C_a$ coordinate of CDR residues \cite{jin2021iterative}.

\subsection{Compared Methods}
\begin{itemize}
    \item \textbf{RosettaAntibody-G}\footnote{\url{https://rosie.rosettacommons.org/}} uses a combination of energy-based methods and sampling techniques to assemble complete Fv structures from sequence-similar fragments of previously solved structures \cite{jeliazkov2021robustification,weitzner2017modeling}.
    \item \textbf{ABodyBuilder}\footnote{\url{https://opig.stats.ox.ac.uk/webapps/newsabdab/sabpred/abodybuilder2/}} employs a combination of computational methods, including an iterative approach to homology modeling and loop modeling \cite{leem2016abodybuilder}.
    \item \textbf{DeepAb}\footnote{\url{https://github.com/RosettaCommons/DeepAb}} determines the geometries between residues, which are then used to form the final structure via an energy minimization technique \cite{ruffolo2022antibody}.
    \item \textbf{ABlooper}\footnote{\url{https://github.com/oxpig/ABlooper}} uses equivariant graph neural networks to predict the structures \cite{abanades2022ablooper}.
    \item \textbf{RefineGNN}\footnote{\url{https://github.com/wengong-jin/RefineGNN}} co-designs the sequence and 3D structure as graphs. The model unravels the amino acid sequence and iteratively refines its predicted structure \cite{jin2021iterative}.
    \item \textbf{MLSA-no att} is a variant of the proposed model without the loop specific attention layer.
    \item \textbf{MLSA-att} is a variant of the proposed model, by replacing the loop specific attention layer with a conventional multi-head attention mechanism.
    \item \textbf{MLSA}\footnote{\url{https://anonymous.4open.science/r/MLSA-CE75/}} is the proposed model.
\end{itemize}

For all baselines, we used the publicly available implementation and kept the parameter tuning as reported, provided that the same datasets were used. For the rest of baselines, the proposed MLSA model and its variants we tuned the parameters via cross-validation and report the best average RMSD. The loop specific attention layer consists of 2 convolutional layers, each of which has a kernel size of [2 × 2]. Both MPNs have 4 message passing layers with hidden size 256. As for the optimization, we use the Adam optimizer with a 0.001 learning rate.

\subsection{Performance Evaluation}
In Table~\ref{tab:perf}, we present the experimental results, when comparing the examined models in terms of average RMSD in the challenging case of the $H_3$ loop. \footnote{Regarding the other two loops, the average RMSDs are 
$H_1$:~$\{1.15, 0.65, 0.74, 0.63, 0.59, 0.72, 0.71, 0.69\}$  and 
$H_2$:~$\{1.00, 0.61, 0.82, 0.74, 0.82, 1.31, 0.87, 0.81\}$ for RosettaAntibody-G, ABodyBuilder, DeepAb, ABlooper, RefineGNN, MLSA-no att, MLSA-att and MLSA, respectively.} We also report the relative improvement of the proposed MLSA method, when compared with its variants and the baselines. On inspection of Table~\ref{tab:perf}, we observe that our models, that is MLSA-no att, MLSA-att and MLSA reduce the average RMSD, when compared with the other baselines for the $H_3$ loop. This indicates that the joint learning of the three CDR loops can significantly increase the prediction accuracy for the $H_3$ loop. The proposed MLSA model outperforms the best method of the baselines, that is RefineGNN by 19\%. In addition, the experimental results show the importance of the attention mechanism. For example, when evaluating MLSA-no att against MLSA-att and MLSA, it is clear that the absence of an attention mechanism can introduce noise to the joint learning strategy. MLSA, achieves higher performance than MLSA-att, by capturing more accurate the functional and structural similarities and differences of the three CDR loops, via the loop specific attention layer. 

\begin{table}
\caption{Average RMSD and relative improvement of the proposed MLSA method, when compared with its variants and the baselines, for the $H_3$ loop.}
\centering
\begin{tabular}{| c | c | c |}
\hline
Methods & $H_3$ RMSD & MLSA Improv.$\%$  \\
\hline
RosettaAntibody-G & $2.7037 $ & $32.843\%$ \\
\hline
ABodyBuilder & $2.4597$ & $26.182\%$ \\
\hline
DeepAb & $2.4110$ & $24.691\%$ \\
\hline
ABlooper & $2.2737$ & $20.143\%$ \\
\hline
RefineGNN & $2.2510$ & $19.338\%$ \\
\hline\hline
MLSA-no att & $2.2002$ & $17.475\%$ \\
\hline
MLSA-att & $1.9378$ & $6.301\%$ \\
\hline
MLSA & $\textbf{1.8157}$ & $0.000\%$ \\
\hline
\end{tabular} \label{tab:perf}
\end{table}

\begin{table*}[t]
\caption{Average RMSD of the examined models for different loop lengths of $H_3$. Bold values indicate the best method.}
\centering
\resizebox{\textwidth}{!}{\begin{tabular}{| c | c | c | c | c | c | c | c | c | c | c | c |} \hline
& \multicolumn{11}{|c|}{Loop length} \\ \cline{2-12}
Methods & 7 & 8 & 9 & 10 & 11 & 12 & 13 & 14 & 15 & 16 & 17 \\
\hline
RosettaAntibody-G & 2.5097 & 2.0260 & \textbf{1.0163} & 2.7082 & 1.7320 & 2.3615 & 4.9170 & 4.4917 & 2.7056 & 4.7823 & 6.9455 \\
\hline
ABodyBuilder & 1.4806 & 2.5513 & 1.7407 & 2.7977 & 1.7235 & 2.7025 & 2.0050 & 2.8800 & 3.7730 & 3.4937 & 4.2395 \\
\hline
DeepAb & 1.5046 & 2.6790 & 1.5313 & 2.4651 & 1.9657 & 2.6438 & 2.0320 & 3.1348 & 3.7350 & 3.7953 & 4.1595 \\
\hline
ABlooper & 1.8936 & \textbf{1.0454} & 1.7298 & 2.1618 & 1.7628 & 2.5535 & 1.9622 & 2.7876 & 3.8116 & 3.2749 & 4.0924 \\
\hline
RefineGNN & 1.8404 & 1.7496 & 1.9610 & 2.1535 & 1.9574 & 2.3198 & 2.1294 & 2.8607 & 2.5203 & 3.5254 & \textbf{2.8529} \\
\hline\hline
MLSA-no att & 1.4161 & 2.0591 & 1.6205 & 1.9242 & 2.1127 & 2.2227 & 1.8343 & 2.6637 & 3.0116 & 3.1681 & 3.5340 \\
\hline
MLSA-att & 1.2671 & 1.7816 & 1.3142 & 1.8746 & 1.8762 & 2.0532 & 1.7402 & 2.4499 & 2.0256 & 3.1686 & 3.2744 \\
\hline
MLSA & \textbf{1.1442} & 1.4993 & 1.1280 & \textbf{1.7353} & \textbf{1.6014} & \textbf{1.9096} & \textbf{1.5899} & \textbf{2.4310} & \textbf{1.9734} & \textbf{2.9320} & 3.3954 \\
\hline
\end{tabular}} \label{tab:length}
\end{table*}

As aforementioned in Section~\ref{chap1}, the loop length of $H_3$ varies. In particular, for our evaluation datasets the loop length varies from 7 to 17 by a step of 1. In Table~\ref{tab:length}, we report the performance of each examined model for the 11 different loop lengths. MLSA clearly outperforms the baselines, by achieving the lowest RMSD for 8 out of the 11 loop lengths. This means, that the proposed MLSA model is accurate for different loop lengths, which is important in the challenging case of the $H_3$ loop.

\begin{figure}
    \centering
    \begin{subfigure}[b]{0.24\textwidth}
        \centering
        \includegraphics[width=0.8\textwidth]{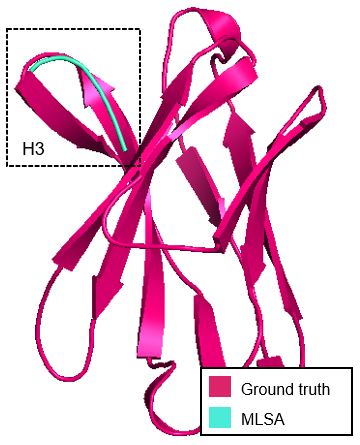}
        \caption{}
        \label{Visualization1}
    \end{subfigure}
    \hfill
    \begin{subfigure}[b]{0.24\textwidth}
        \centering
        \includegraphics[width=0.8\textwidth]{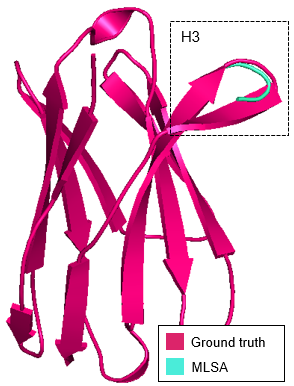}
        \caption{}
        \label{Visualization2}
    \end{subfigure}
    \caption{Visualization of an antibody example of a ground truth $H_3$ loop structure and
    the predicted structure by MLSA. Figures \ref{Visualization1} and \ref{Visualization2} refer to the same antibody example with a $180\degree$ rotation.}
    \label{Visualization}
\end{figure}

In Figure \ref{Visualization} (PDB: 1nlb, RMSD: 0.7459), we visualize an antibody and the corresponding predicted $H_3$ loop , using PyMOL\footnote{\url{https://pymol.org/2/}}, an open-source molecular visualization system. Figures~\ref{Visualization1} and~\ref{Visualization2} refer to the same representation with a $180\degree$ rotation. The arrows show the direction of the antigen-binding sites, that is the regions of the antibody that bind to antigens.

\subsection{Loop Specific Attention Tuning}

\begin{figure}
    \centering
    \begin{subfigure}[b]{0.24\textwidth}
        \centering
        \includegraphics[width=\textwidth]{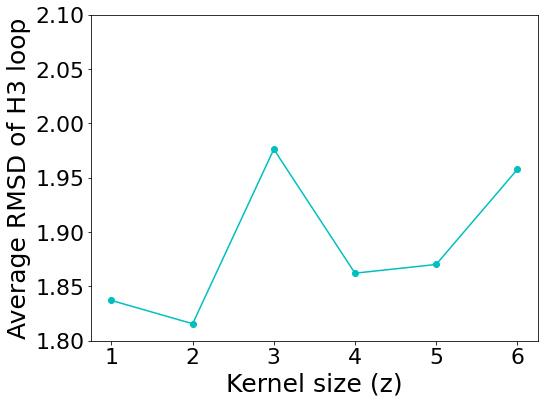}
        \caption{}
        \label{tuning_kernel}
    \end{subfigure}
    \hfill
    \begin{subfigure}[b]{0.24\textwidth}
        \centering
        \includegraphics[width=\textwidth]{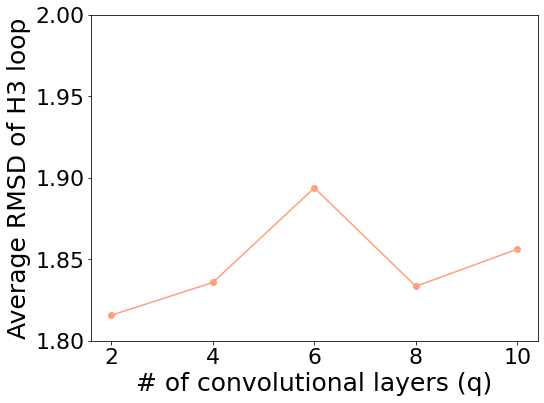}
        \caption{}
        \label{tuning_layers}
    \end{subfigure}
    \caption{Average RMSD of MLSA when varying (a) the kernel size $z$ of each convolutional layer and (b) the \# of convolutional layers $q$ for the $H_3$ loop. }
    \label{Visualization}
\end{figure}

The loop specific attention mechanism of MLSA plays a crucial role in the prediction accuracy of the proposed MLSA model. Two are the most important parameters in the loop specific attention mechanism, that is the kernel size $z$ of each convolutional layer and the number of convolutional layers $q$. In Figures~\ref{tuning_kernel} and~\ref{tuning_layers} we report average RMSD when varying the kernel size and the number of convolutional layers. Provided that the training cost significantly increases for high kernel sizes and numbers of convolutional layers we vary  $z\in$~[1~6] and $q\in$~[2~10] by a step of 1 and 2, respectively. As presented in Figures~\ref{tuning_kernel} and~\ref{tuning_layers} we fix $z$ and $q$ to 2 and 2, accordingly, as higher values of both parameters do not reduce the average RMSD for the $H_3$ loop. 

\subsection{Discussion}

The task of the proposed MLSA method, that is to predict the structures of CDRs, is significant as it is a valuable tool for important advances in healthcare and medicine. Antibodies have been already used successfully to prevent a wide range of diseases, like cancer, transplant rejection, asthma, Crohn’s disease, and infectious diseases \cite{lu2020development}. Although considerable progress has been made in discovering the interactions between antigens and host immune systems, there are still many unresolved questions. By better understanding how antibodies recognize and bind to antigens via CDR structure prediction, researchers can develop more effective therapeutics to treat a variety of diseases, such as vaccines and antibody-based drugs, that target specific antigens with high precision.  

In addition, the accurate prediction of the CDR structures is crucial for antibody engineering. In particular, antibody-based drugs are a type of biologic therapy that has revolutionized the treatment of many diseases and represent the fastest growing class of pharmaceuticals on the market \cite{carter2018next}. They have several advantages over conventional drugs, including high specificity and potency, low toxicity, and metabolic stability. Antibody-based drugs have been used to treat several diseases, including cancer, autoimmune disorders, infectious diseases, and cardiovascular diseases. According to \cite{kaplon2023antibodies}, nearly 1,200 antibody therapeutics are currently in clinical studies and approximately 175 are in regulatory review or approved. Ongoing research in this field holds promise for the generation of new and more effective treatments for many diseases, making the accurate prediction of CDR structures important in antibody engineering.

Globally, infectious diseases caused by viruses, bacteria, and other pathogens continue to be of great concern for public health. The prediction of CDR structures for antibodies offer key advantages for the treatment and prevention of infectious diseases, as they can target and neutralize pathogens by binding to specific antigens on their surface, preventing them from infecting host cells \cite{wagner2018engineering}. Numerous antibodies have shown great potential in the treatment of infectious diseases in preclinical testings. Vitro studies have demonstrated the ability of antibodies to destroy pathogens and stop them from invading host cells. In animal models, antibodies have been shown to reduce the severity and duration of infections, and even prevent infection altogether \cite{chan2009use}. In addition, antibodies and antibiotics are often used in combination for the treatment of infectious diseases. When used together, antibodies and antibiotics can exhibit additive or synergistic effects, leading to more efficient bacterial clearance and improved patient outcomes \cite{wagner2018engineering}. 

Cancer is one of the major challenges in healthcare. Conventional cancer treatments, such as chemotherapy and radiation, suffer from several drawbacks, including lack of specificity, toxicity to healthy cells, and development of drug resistance. In addition, these therapies often fail to eliminate all cancer cells, leading to disease recurrence. In recent years, antibody-based drugs have emerged as a promising class of therapeutics for the treatment of cancer, accounting for almost half of all marketed antibody therapeutics \cite{chen2022antibody}. These drugs work by targeting specific proteins or cells that are overexpressed or abnormally expressed in cancer cells, while sparing normal cells. Monoclonal antibodies are the most commonly used type of antibody-based drugs in cancer therapy, as they can deliver the drugs to the selective antigens \cite{sapra2013monoclonal}.

Antibody-based medications have also been studied as potential treatments for HIV, the virus that causes AIDS. Therapeutic antibodies have become a promising class of drugs for HIV treatment. These antibodies target various components of the HIV virus, such as the envelope protein or the CD4 receptor on immune cells \cite{lu2020development}. One such antibody, called VRC01, has shown promising results in clinical trials, neutralizing a wide range of HIV strains and reducing viral load in patients \cite{zhou2013multidonor}. In addition to their direct antiviral effects, therapeutic antibodies also have the potential to stimulate the immune system and enhance antiviral responses. Although the development of therapeutic antibodies for HIV is still at an early stage, they hold promise for improving HIV treatment and prevention in the future.

Furthermore, there are several studies that demonstrate the potential of personalized antibody-based therapies \cite{ott2017immunogenic}. CDR structure prediction can contribute to the field of personalized medicine by enabling the development of customized antibody-based therapies. Personalized medicine involves tailoring medical care to the unique characteristics of each patient, such as their genetic makeup, disease status, and response to treatment. Antibody-based therapies are attractive candidates for personalized medicine, due to their specificity and the ability for tailoring them to the individual patient's needs. CDR structure prediction can be used to design antibodies with specific binding properties to target individual disease-related antigens or even patient-specific antigens. This way, CDR structure prediction can facilitate the development of personalized antibody-based therapies that could improve treatment outcomes and reduce the risk of adverse reactions.

\section{Conclusion}
\label{chap5}
In this study we presented the MLSA model for CDR 3D structure prediction. The key idea is to perform joint learning of the CDR loops of the heavy chain. A crucial role to boost the prediction accuracy is the loop specific attention mechanism, which balances the impact of each CDR loop on the joint learning of MLSA. Our experimental evaluation demonstrated that the proposed MLSA model significantly reduces the prediction error by 19\% in the challenging case of the $H_3$ loop, when compared with the best state-of-the-art strategy. An interesting future direction is to extend the MLSA model performing MTL not only on different CDRs from the heavy chain, that is the $H_1$, $H_2$ and $H_3$ loops, but also from the light chain, that is the $L_1$, $L_2$ and $L_3$ loops, as the loops from both chains have a similar structure and are involved in antigen binding. However, performing MTL on both chains is a challenging task, as the heavy chain is typically larger and contains more domains than the light chain \cite{chiu2019antibody,stanfield2015antibody}.

\bibliographystyle{ieeetr}

\end{document}